\documentclass[10pt, journal, letterpaper]{IEEEtran}

\usepackage{url}

\usepackage{color}

\usepackage{flushend}

\usepackage{cite}
\usepackage{amssymb}

\ifCLASSINFOpdf
   \usepackage[pdftex]{graphicx}
\else
   \usepackage[dvips]{graphicx}
\fi
\usepackage{subfigure}
\usepackage{multirow}
\usepackage{float}

\usepackage[cmex10]{amsmath}

\usepackage{array}

\usepackage{balance}

\usepackage{algorithm}
\usepackage{algpseudocode}

\begin{document}
%
\title{Transforming Cooling Optimization for Green Data Center via Deep Reinforcement Learning}
\author{Yuanlong Li,
        Yonggang Wen,
        Kyle Guan,
        and Dacheng Tao

\thanks{Yuanlong Li and Yonggang Wen are with School of Computer Science and Engineering, Nanyang Technological University, Nanyang Avenue, Singapore 639798. Email: \{liyuanl, ygwen\}@ntu.edu.sg.}

\thanks{K. Guan is with Bell Labs, Nokia, Holmdel, NJ 07733 USA.}

\thanks{Dacheng Tao is with the Centre for Quantum Computation and Intelligent Systems and the Faculty of Engineering and Information Technology, University of Technology, Sydney, 81 Broadway Street, Ultimo, NSW, Australia.}
}

\maketitle

\begin{abstract}
Cooling system plays a critical role in a modern data center (DC). Developing an optimal control policy for DC cooling system is a challenging task. The prevailing approaches often rely on approximating system models that are built upon the knowledge of mechanical cooling, electrical and thermal management, which is difficult to design and may lead to sub-optimal or unstable performances. In this paper, we propose utilizing the large amount of monitoring data in DC to optimize the control policy. To do so, we cast the cooling control policy design into an energy cost minimization problem with temperature constraints, and tap it into the emerging deep reinforcement learning (DRL) framework. Specifically, we propose an end-to-end cooling control algorithm (CCA) that is based on the actor-critic framework and an off-policy offline version of the deep deterministic policy gradient (DDPG) algorithm. In the proposed CCA, an evaluation network is trained to predict an energy cost counter penalized by the cooling status of the DC room, and a policy network is trained to predict optimized control settings when given the current load and weather information.
The proposed algorithm is evaluated on the EnergyPlus simulation platform and on a real data trace collected from the National Super Computing Centre (NSCC) of Singapore. Our results show that the proposed CCA can achieve about 11\% cooling cost saving on the simulation platform compared with a manually configured baseline control algorithm. In the trace-based study, we propose a de-underestimation validation mechanism as we cannot directly test the algorithm on a real DC. Even though with DUE the results are conservative, we can still achieve about 15\% cooling energy saving on the NSCC data trace if we set the inlet temperature threshold at 26.6 degree Celsius.

\end{abstract}

\begin{IEEEkeywords}
Deep learning, reinforcement learning, data center cooling optimization.
\end{IEEEkeywords}


\IEEEpeerreviewmaketitle

\section{Introduction}

With the emergence and proliferation of services and applications such as cloud computing and social networks, data center (DC) plays an ever important role. It is predicted that global DC IP traffic will grow 3-fold from 2014 to 2019 with a compound annual growth rate (CAGR) of 25 percent \cite{CiscoGCI}. At the same time, the high energy consumption of DCs is drawing more and more attention due to economic, social, and environmental concerns. DC electricity consumption in the U.S. alone is projected to increase to roughly 140 billion kilowatt-hours annually by 2020, costing \$13 billion in electricity bills and emitting nearly 100 million tons of carbon pollution per year~\cite{DataCentreEnergyGrow}. In this paper, we focus on one of the significant sources of energy consumption in DC (about 38\%~\cite{ni2017review}), the cooling energy.

Cooling energy optimization involves the control of a sophisticated cooling system, which consists of multiple components, such as cooling tower, chiller, and ventilation system, etc. A common practice of DC cooling system control is to adjust the set-points, i.e., the target values of different control variables. For example, by setting the temperature control variable at the outlet of an air conditioner to the desired value, the air conditioner can adjust its internal state to meet the set-point by consuming a certain amount of energy. An optimal selection of these set-points can be challenging, as the process relies on the knowledge of the cooling system, from thermal dynamics to mechanics. Many existing approaches are based on an approximated system model that often incorporates the first-order effects of thermal, electrical, and mechanical principles \cite{sun2005optimal, ahn2001optimal,lu2005hvac,chang2004novel,ma2009optimal}. These approximated models are sometimes either inadequate or inaccurate to capture the intricacies of various interacting processes of DC cooling operations, leading to sub-optimal or unstable cooling controls. Recently, the learning-based approach has emerged as an attractive alternative. A learning-based approach does not assume any specific model of the underlying system. Instead, the control policies are learned and derived from the massive data collected on the system status and energy consumption \cite{chow2002global}. This approach is especially advantageous when the complexity of the underlying system makes an accurate mathematically modeling a daunting task.

Currently, the general cooling control optimization approach, which includes a model building stage and a solving stage, can be referred to as the two-stage (TS) approach. In comparison, an end-to-end or one-stage approach uses directly the unprocessed and often high-dimensional input to learn a control policy, which can be used to determine the control setting given the system input. One such framework to our interest is reinforcement learning (RL) \cite{sutton1998reinforcement}, in which neural based control agent has been introduced decades ago \cite{barto1983neuronlike}. Recently algorithms that combine RL with deep learning (DL), such as deep Q-network (DQN), have been successfully applied to the task of training AI agent to play video games at the human performance level with only raw pixel inputs, thus showing the potential of the end-to-end approach \cite{mnih2013playing}. The continuous domain extension of DQN, called deep deterministic policy gradient (DDPG), has also shown promising results on simulated physical control tasks \cite{lillicrap2015continuous}. However, DDPG has not been widely studied in the context of a more realistic and complex control optimization, such as cooling system optimization for DCs (the subject of this paper). It remains to be demonstrated if an end-to-end approach can achieve similar or better control performance compared with the TS approaches. Besides, DDPG is a simulation-based algorithm. Like many RL algorithms, it uses an excessive amount of computation and possibly very long training time. Thus it is still interesting to see how the challenges are met. Note that Google claims that they use AI method \cite{googlePUE} to reduce the PUE of their DC yet no detailed methodology or performance evaluation results are disclosed. In our previous work, we reviewed the DC energy cost models \cite{dayarathna2016data} and the existed cooling optimization approaches \cite{zhang2016towards}, and conducted several DC power analysis and control studies \cite{xia2015toward} \cite{li2016learning} \cite{yin2013cloud3dview}; based on which we believe that a data-driven learning-based optimization method is needed for DC energy optimization, which can be used to achieve optimization effects with minimum human innervations and reduce the DC management difficulty.

In this paper, we propose an end-to-end approach for DC cooling control optimization and evaluate the algorithm from various aspects. We develop a cooling control algorithm adapted from the DDPG and the actor-critic architecture \cite{lillicrap2015continuous} \cite{grondman2012survey}. Our proposed algorithm is off-policy and offline, as it can be trained with a pre-collected trace to learn and improve the control policy. Besides the standard version of the algorithm which makes control decisions based on the current state, we also test the recurrent version of the algorithm which can perform better when the data are noisy. We also evaluate different algorithm implementation details such as neural network architecture and hyper-parameters to examine the approach thoroughly.

To test the proposed algorithm, we use the EnergyPlus \cite{crawley2001energyplus} platform to build a test case; besides the simulation case, we also collect a real data trace from the National Super Computing Centre (NSCC) of Singapore and test our algorithm on it. For the simulation test case, we control five different set-points to achieve minimum PUE and to maintain the temperature of the DC zone within a pre-defined range. The results of the proposed algorithms are compared with those generated by a standard two-stage control algorithm and the default set-point based control algorithm (embedded in the simulation software). The results indicate that the proposed algorithm not only successfully maintains the temperature of the DC zone within a pre-defined range under varying workload and weather conditions, but also achieves lower PUE and save about 11\% cooling cost compared with the baseline algorithm. For the real data from NSCC test case, we focus on optimizing the airflow rate setting of the three precision cooling units (PCUs) which are used to cool 26 racks. Our results show that the proposed algorithm can approximate the actual temperature with high accuracy (lower than 0.1 degrees) and can output control settings according to the cooling requirements with around 15\% energy saving. The main contributions of the paper are summarized as follows.

First, we propose an end-to-end and DRL-based framework that can be utilized for DC cooling control optimization. We propose an algorithm that trains the neural network with a pre-collected data trace, such that we can overcome the problem of high simulation time cost in simulation-based algorithms like DDPG. This approach is well suited for a practical DC equipped with a monitoring/sensing system that collects data in real time.

Second, we build a test-bed with EnergyPlus and verify the approach based on the sophisticated simulation software. Our simulation results indicate that the proposed control algorithm can accomplish the cooling control tasks with about 11\% cooling cost saving compared with the baseline approach.

Third, we propose a de-underestimate (DUE) solution for trace-based study in practice. The DUE method can be used to eliminate underestimation of the predicted temperature and thus drives to a more conservative and low-risk energy saving calculation when a real test or decent simulation is not available. Such a method is useful as risk management is essential for DC operation.

In summary, we demonstrate the feasibility and effectiveness of applying an end-to-end neural control algorithm to the DC cooling optimization. The evaluation of the performance of the proposed algorithm serves as the first step to build an intelligent DC management system that requires minimal manual intervention. Though this work is simulation and trace based, it does shed new light on the application of deep reinforcement learning (DRL) to practical DC control optimization, and application of DRL to other traditional industry areas.

\section{Related Works}
\label{stn:related}

\subsection{Recent Progress on DC Cooling Optimization}
Cooling system control optimization problem has been examined from different aspects. A lot of these literatures \cite{sun2005optimal} \cite{ahn2001optimal} \cite{lu2005hvac} \cite{chang2004novel} \cite{ma2009optimal} \cite{chow2002global} \cite{fouladi2017optimization} focus on using a two-stage optimization procedure to optimize the cooling efficiency. For example, in the first stage, a thermal dynamics model is built to evaluate the efficiency, such as in \cite{sun2005optimal} \cite{ahn2001optimal} \cite{lu2005hvac} \cite{chang2004novel} \cite{ma2009optimal} \cite{fouladi2017optimization}. Recently in \cite{singh2018system}, the authors proposed utilizing computational fluid dynamics (CFD) model to analyze the airflow efficiency. These models can be very complex, such as in \cite{sun2005optimal} the authors proposed a mathematical model for a specific cooling system which including 43 equations. With such complexity, this kind of model is hard to extrapolate to another DC with a separate configuration. Research has been trying to build data driven approach, such as the neural network model in \cite{chow2002global}.
In the second stage, the control variables are optimized via either an analytic optimization algorithm such as in \cite{sun2005optimal} or a random global optimizer such as in \cite{chow2002global}. Different to the existing TS approach, we propose to train a policy network in an offline manner such that we can avoid the optimization procedure during the decision making process.

Another area of research on cooling control optimization is to optimize the ice-storage system \cite{braun1990reducing} \cite{lo2016ice}, so that it can be used for cooling when the electricity price is high.
In addition to directly optimizing the cooling system surveyed above, there exists an extensive body of works that focus on the IT side of the DC to save cooling energy. For example, in \cite{pakbaznia2009minimizing} \cite{li2014coordinating} \cite{banerjee2011integrating}, the workload dispatch problem was studied to optimize the thermal map in DC to improve cooling efficiency. In \cite{zheng2017powernets}, the authors proposed a novel network flow management method to use fewer switches to save power. Our approach can be combined with IT side optimization to reduce the energy cost of DC further.

We note that most of the existing studies are based on ideal models in a simplified situation. In this paper, we propose a new DRL based solution and verify the approach on both complicated simulation system and a real data trace. 

\subsection{Recent Progress on DRL}
Reinforcement Learning \cite{barto1983neuronlike} \cite{lewis2013reinforcement} deals with agents that learn to take better actions directly from experience of interacting with the environment.
Recently the development and application of RL technologies have flourished. For example, in \cite{liu2014policy}, Liu et al. proposed an adaptive dynamic programming method to do policy iteration for nonlinear systems;
in \cite{luo2015off}, Luo et al. proposed a neural actor-critic RL solution to the  $H_\infty$ control problem;
in \cite{liu2015reinforcement}, Liu et al. proposed a single critic network based RL solution to the constrained-input stabilizing controller;
in \cite{modares2016optimized}, Modares et al. applied RL to the human-robot interaction system which can minimize the human effort and optimize the control results;
in \cite{pan2017biomimetic}, Pan et al. proposed a neural control algorithm that mimics the human motor learning; in \cite{song2017off}. Song et al. propose an off-policy RL method to solve nonlinear nonzero-sum games; in \cite{deng2017deep}, Deng et al. build a financial trading agent based on deep neural networks.

In the flourish of RL studies and applications, deep reinforcement learning (DRL) \cite{mnih2013playing} has shown its strength in various fields.
The deep Q-network (DQN) proposed in \cite{mnih2013playing} applies a neural network approximation to the Q table in Q-learning \cite{watkins1992q}. Subsequent studies on DQN have been focusing on improving the training stability of the framework such as in \cite{van2015deep} and extending the framework to solve problems with continuous control variables \cite{lillicrap2015continuous}. Various applications of these deep reinforcement learning have been proposed such as video processing \cite{koutnik2014evolving} and text-based game \cite{narasimhan2015language}. Yet DRL has not been verified in the practical control system like the cooling system in a DC, where high simulation cost can be trouble, and the robustness requirement is high. In this paper, we propose to adopt the DDPG algorithm for the cooling system control optimization problem and examine various implementation details related to the robustness of the algorithm.

\section{The Cooling Optimization Problem Formulation} \label{modelandform}
To study the cooling control optimization problem, in this section we utilize a simulation model to present a cooling control optimization problem formulation. The simulation model is based on the widely adopted building energy simulation platform EnergyPlus \cite{crawley2001energyplus}. Although the model is largely simplified, it does capture the major cooling dynamics and is thus adequate for studying the cooling control optimization.

\subsection{Simulation System Model} \label{model}

\begin{figure}[]
\centering
   \includegraphics[width=0.5\columnwidth] {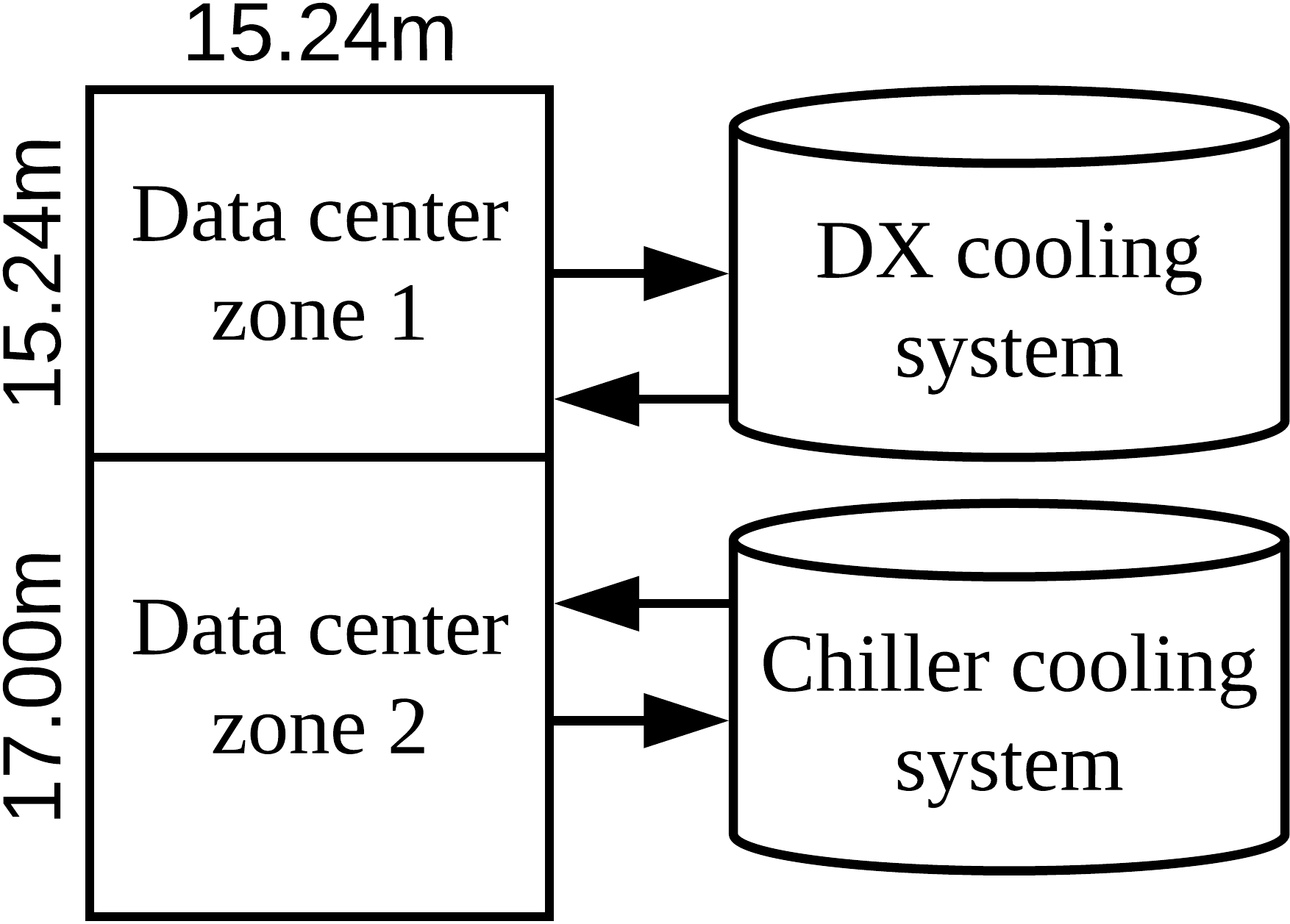}
\caption{System architecture:  a DC consists of two server zones, $z_1$ and $z_2$, and their associated and independently operated cooling facilities. For $z_1$ direct expansion (DX) cooling system is used and for $z_2$ chiller plant cooling system is used.}

\label{fig:sysarc}
\end{figure}

The model is based on a simulation example provided by EnergyPlus. As illustrated in Fig. \ref{fig:sysarc}, the model consists of two server zones ($z_1$ and $z_2$) and their associated cooling systems. The two zones are different in size, location, and their corresponding cooling systems. In the following, we describe the cooling system with a focus on identifying the state space (parameters that characterize the system), the action space (control variables), and the reward counters (optimization objectives), while omitting the details of the facility structures and operation processes that are unrelated to problem setup (yet might be critical parts of the overall system).

\begin{figure}[]
\centering
   \includegraphics[width=0.7\columnwidth] {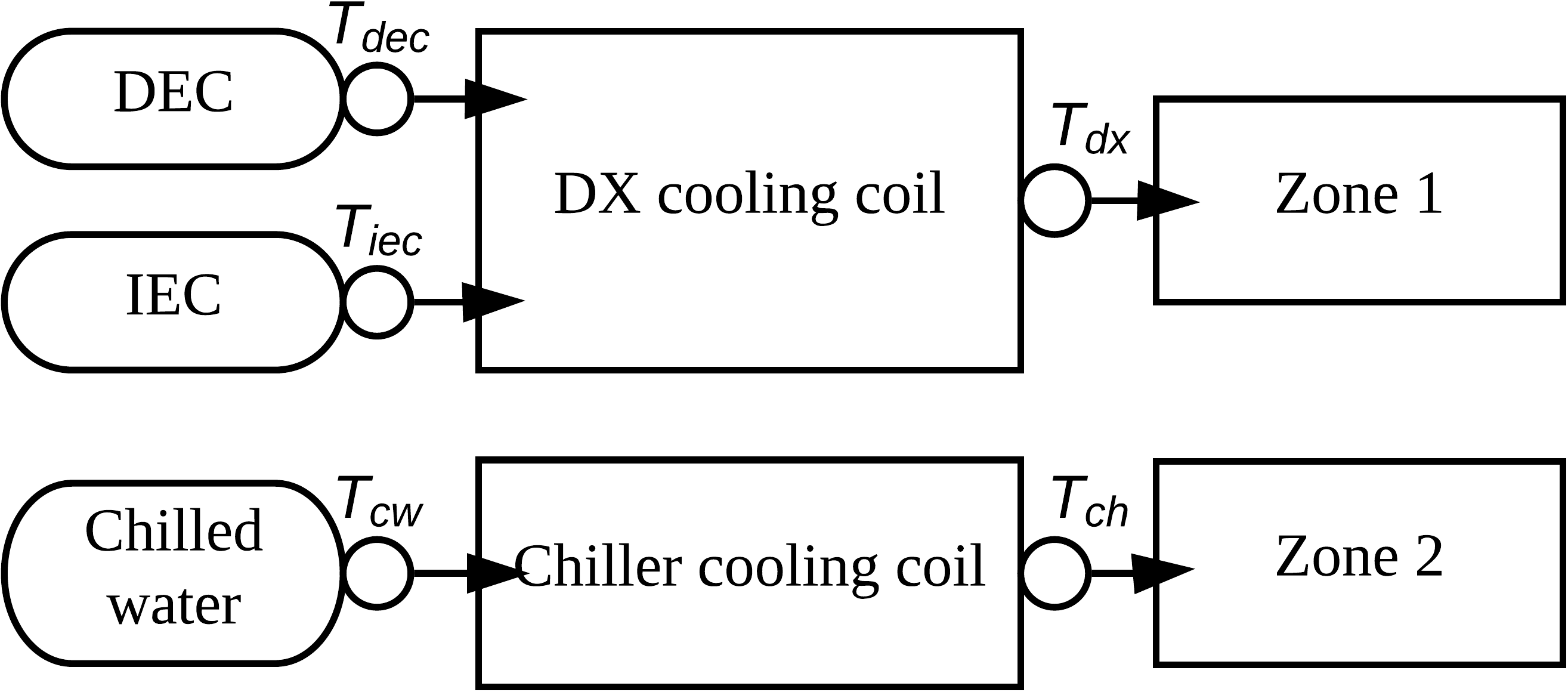}   
\caption{The main components of DX and chiller cooling system. The temperatures at five outlet ports (marked as open circles) are served as the temperature set-points: $T_{dec}$, $T_{iec}$, $T_{cw}$, $T_{dx}$, and $T_{ch}$.}

\label{fig:coolarc}
\end{figure}

\subsubsection{Data Center Model -- State Space and Reward}
The DC has two server zones placed side by side, $z_1$ and $z_2$, with each zone being a standalone server room. The two zones are different in size ( 15.24*15.24$m^2$  and 15.24*17.00$m^2$ for $z_1$ and $z_2$, respectively) but similar in other construction aspects. The heat in each zone is generated by IT equipment (ITE) and other sources (such as illuminations), with ITE as the dominant heat source. The load of the ITE is defined as $\alpha \cdot L$, where $L$ is the designed load density (per square meter) and $\alpha$ is a load factor that varies at different time slots. In our simulation, we use a public trace collected from Wikimedia \cite{wikitrace} to set $\alpha$ to be the same for $z_1$ and $z_2$ while using different load density 4kw and 2kw for $z_1$ and $z_2$ respectively. The heat generated by the illumination is assumed to be a constant (per square meter) as the lights inside the DC are on all the time. Also, there is heat generated by the human workers in the DC. This part varies according to the work schedule. A notable simplification of the simulation is that each data zone is modeled as a single point-heat source. This is less accurate compared to the CFD based thermal analysis. We leave a finer-grained model for future work. We note that even with a finer-grained model, the proposed framework and algorithm will remain the same, albeit with much larger state space and action space.

In the context of the RL framework, we use a tuple of workload level and the ambient temperature to represent the state, since both of them affect the cooling load. We use the tuple of PUE and the IT equipment outlet temperature of each zone to compute the reward. In the context of the DC cooling, PUE needs to be minimized and the outlet temperature needs to be kept within a specific range.

\subsubsection{Cooling System Model: Action Space}
In the target simulation model, $z_1$ and $z_2$ are equipped with different cooling systems: direct expansion (DX) cooling system for $z_1$ and chilled water (chiller) cooling system for $z_2$. Both cooling systems are supplied with cool water from a cooling tower, but they use the cool water in different ways. In the DX system, the cool water passes through coils and cool down the airflow passing over the coils. In the chiller system, the cool water is used first to refrigerate another water stream (chilled water), which in turn cools down the airflow supplied to the DC.

The main components of these two cooling systems are shown in Fig. \ref{fig:coolarc}. In the DX cooling system, the intake ambient air flow is first cooled by two types of evaporative coolers: directive (DEC) and indirect (IEC), and then passes over the DX cooling coils and is further fed to the DC.

The underlying control algorithm in EnergyPlus (referred as the DefaultE+ control algorithm) uses the following five set-points to control the cooling system:
DEC outlet temperature (airflow) $T_{dec}$,
IEC outlet temperature (airflow) $T_{iec}$,
chilled water loop outlet temperature (water flow) $T_{cw}$,
DX cooling coil outlet temperature (airflow) $T_{dx}$,
and chiller cooling air loop outlet temperature (airflow) $T_{ch}$.

The DefaultE+ algorithm relies on a fixed zone temperature set-point setting, and then compute the settings of the five set-points based on the knowledge of the underlying system dynamics with the load and weather information. In our proposed learning algorithm, the same five set-points are used as control variables, but, on the contrary, they are learned from the pre-collected data trace. Neither physical meaning of these variables nor the relationship information among them will be used in training.

\subsection{Problem Statement} \label{formulation}
We formulate the cooling control optimization problem as follows. We are given a time-varying tuple of the ambient air temperature $T_{amb}$ and the load factor $H_{ite}$. The problem is to determine the values of five control set-points to minimize the objective function as stated in Eq. \eqref{eq:opt}.

\newcounter{MYtempeqncnt}
\begin{figure*}[t!]
\normalsize
\setcounter{MYtempeqncnt}{\value{equation}}
\setcounter{equation}{0}

\begin{eqnarray}
\min_{T_{cw},T_{dx},T_{dec}, T_{iec},T_{ch}}    && \epsilon_{pue} + \lambda \cdot \ln(1+\exp(T_{z_1}-\phi)) + \lambda \cdot \ln(1+\exp(T_{z_2}-\phi)) , \label{eq:opt}\\
\text{s.t.} && L_{T_{cw}} \le T_{cw} \le U_{T_{cw}}, \nonumber \\
            && L_{T_{dx}} \le T_{dx} \le U_{T_{dx}}, \nonumber \\
            && L_{T_{dec}} \le T_{dec} \le U_{T_{dec}}, \nonumber \\
            && L_{T_{iec}} \le T_{iec} \le U_{T_{iec}}, \nonumber \\
            && L_{T_{ch}} \le T_{ch} \le U_{T_{ch}}. \nonumber
\end{eqnarray}

\setcounter{equation}{\value{MYtempeqncnt}}

\hrulefill

\vspace*{4pt}
\end{figure*}
 The objective function aims to strike a balance between minimizing the PUE and preventing overheating in the server zone. In particular, the objective function consists of two parts: the first part is PUE (denoted as $\epsilon_{pue}$), which is to be minimized; the second part accounts for the penalty of the overheating (for both $z_1$ and $z_2$).  The penalty function has the form of  $\lambda$ $\cdot ln(1+\text{exp}(T_{z_i}-\phi))$, for $i \in [1,2]$, with $\lambda$, $T_{z_i}$, and $\phi$ denoting the penalty pricing factor, average IT equipment outlet temperature of zone $i$, and overheating threshold, respectively. The penalty term takes the standard form of the soft plus activation function which has been implemented in most deep learning frameworks thus easy to be implemented. During the training, we minimize this cost function (reverse of the reward), since commonly the training optimization algorithms are designed for minimization.

\section{The Proposed Approach: Neural End-to-End Cooling Control Algorithm}
In this section, we present the end-to-end cooling control algorithm (CCA), adapted from the DDPG, which combines the critical RL techniques and methods such as deep Q-network (DQN), deterministic policy gradient (DPG), and actor-critic algorithm. In the following, we first provide an overview of the related RL concepts and techniques. We then describe a complete algorithm flow and the design of the neural networks.

\subsection{Overview of Q-learning and Policy Gradient}
For our application, the goal is to enable an AI agent to learn an optimal cooling control policy from a data set that records a sequence of states, actions taken, and rewards at discrete time steps. Within the RL framework, this goal is achieved by using either value-based or policy-based approaches. Central to the value-based approaches is the Q-learning technique. Though for discrete state and action space (especially when space is small), Q-function can be represented as a table computed by Bellman equation iterative updating, in practice it is often estimated by a function approximator such as a neural network, like the Deep Q-network (DQN) \cite{mnih2013playing}. With policy-based approaches, policy-gradient (PG) is an important algorithm that optimizes a policy end-to-end by computing noisy estimates of the gradient of the expected reward and then updating the policy in the gradient direction. When the state or action space is represented by continuous variables, a naive adaptation of DQN or PG via discretization of state or action space often results in intractability or very slow learning convergence (even divergence). We use the DDPG algorithm, which is essentially a hybrid method combining the policy gradient method and the value function \cite{lillicrap2015continuous}.

\subsection{Online Learning vs. Batch Learning and Off-Policy vs. On-Policy}
We note that RL algorithms can be directly used as online learning algorithms. This means that the control algorithm can learn in an online manner, e.g., starting from an initial state and adjusting itself with the input it received from the ongoing process, either the real operation or the simulation. However, this will be problematic for the DC cooling task, which cannot risk erroneous settings. In this work, we focus on the control algorithms that are pre-trained by the offline data first, which is referred to as ``batch learning".  Batch algorithms can be further divided into two categories based on how the training data are generated: off-policy and on-policy. Off-policy algorithms generally employ a separate behavior policy, which is independent of the policy being estimated, to generate the training trace; while on-policy directly uses control policy being estimated (in the real control practice or more likely in a simulator) to generate training data traces. For the case of DC simulation, the cost of simulation time is high. Thus off-policy algorithms are easier to apply and more suitable for our situation.

In summary, we propose an off-policy control algorithm adapted from canonical DDPG. The algorithm employs only a single offline trace for batch learning. In the following, we introduce the details of the proposed algorithm.

\subsection{Cooling Control Algorithm with Offline Trace (CCA)}
The flowchart of the proposed Cooling Control Algorithm (CCA) with an offline trace is shown in Algorithm \ref{alg:CCA}. For the training task, a data trace is collected (line 1), which contains entries: state ($T_{amb}(t)$, $H_{ite}(t)$), action  ($T_{dec}(t)$, $T_{iec}(t)$,$T_{cw}(t)$, $T_{dx}(t)$ and $T_{ch}(t)$), and reward data $y(t)$ computed by the objective function \eqref{eq:opt} based on the observed PUE and temperature data. Different to canonical RL approach in which the future reward data are also included in the evaluation of the action (discounted return), here the future reward information is not used, as the workload and weather trends determine the system transition. Note that as there will be an affecting time for an action to take effect, we shrink the state observation back for one time slot, such that the action we computed based on the current state is going to take effect in the next time slot. All these data are prepared as time series of $N$ time steps, which are further divided into training data and validation data.

Before the training starts, we first initialize two neural networks (line 2). The critic network $Q(X_Q|\theta^Q)$ (parameterized by $\theta^Q$) approximates the Q-value of a state-action pair: it takes the current state and the next action to take (combined into a vector $X_Q$) as the input, and outputs a scalar value which represents the cost of an action $a$ taken at a state $s$. In this paper, we also consider recurrent decision making, which means that a short recent history of states and actions are incorporated in the input of the $Q$ network, i.e., concatenating
 $s_{t-\tau+1}$, $a_{t-\tau+1}$, $s_{t-\tau+2}$, $a_{t-\tau+2}$,..., $s_{t}$, $a_{t}$ into a vector as the input $X_Q$ to the $Q$ network. Recurrent decision making can be helpful when the data are noisy, as will be shown in Section \ref{stn:test_NSCC}. We also propose a special design of the $Q$ network that can ease the training, in which the second last layer of $Q$ is designed to output the predicted PUE and temperature data, with which the cost is computed according to \eqref{eq:opt} in the last layer. With such design, we can easily check the predicted PUE and temperature information from the $Q$ network, which can be helpful to show the quality of the $Q$ network directly. The actor network $\mu(X_\mu|\theta^\mu)$  (parameterized by $\theta^\mu$) is policy network: it takes the recent state-action history and current state ($X_\mu$) and outputs the new action $a_{t}$ to take.

The training procedure is illustrated in lines 3-21. Here we use standard neural network training procedure with multi-training-epochs. Within each epoch (line 5-9), each batch (in random order) of training data is used to update the weights of the neural networks using gradient descent. The critic network is updated by minimizing the mean square error between the output of the second last layer of $Q$ and the raw reward data $y_r$; while the policy network is updated by minimizing the output of $Q$ when taking action at current state according to the output of $\mu$. To avoid over-fitting, we also compute the validation error to keep track of the best weight parameter settings for the two neural networks respectively, as shown in lines 11-20. One important note is that for the $\mu$ network, the validation error can be small in the beginning due to that at that time the $Q$ network is not well learned. For safety, we use a periodical re-initialization of the $E^{\mu}_{val}$ to solve this problem.

\begin{algorithm}[!htb]
\caption{CCA: Cooling Control Algorithm with Offline Trace}
\label{alg:CCA}
\begin{algorithmic}[1]
    \State \emph{Data preparation}: Collect an offline trace with historical data series of the state, action and reward data. Organize the data into the form of $X_Q$, $X_\mu$, and $y_{r}$, where $X_Q$ is the input of the Q-neural network $Q$, $X_{\mu}$ is the input of the policy network $\mu$, and $y_{r}$ is the energy(PUE) and temperature readings that can be used to compute the loss data $y$ according to \eqref{eq:opt}. Divide the data into training data (used for training the neural network) and validation data (used for validating the neural network to find the best parameter settings). Denote $I_V$ as the index of sample data in the validation set.
    \State  \emph{Initialization}: Create neural networks $Q(X_Q|\theta^Q)$ and $\mu(X_\mu|\theta^\mu)$, and randomly set the values of the weight parameters $\theta^Q$ and $\theta^\mu$. Set initial validation error $E^Q_{val}=1e100$, $E^{\mu}_{val}=1e100$ and $Q_{best}=Null$, $\mu_{best}=Null$.
    \For {$Epoch=1,...,MaxEpoch$}
    	\State Randomly divide the training data into batches of size $M$ and denote the index set of the data in the $i^{th}$ batch as $I_i$.
        \For {$i=1,...,N/M$}
            \State Update $\theta^Q$ by minimizing:
            \State $\sum_{j \in I_i}{(y_{r}(j)-Q_{-2}(X_Q(j)|\theta^Q)^2}/M$, where $Q_{-2}$ means the outputs of the second last layer of $Q$.
            \State Update $\theta^\mu$ by minimizing:
            \State $\sum_{j \in I_i}{Q([X_\mu(j),\mu(X_\mu(j)|\theta^\mu)]|\theta^Q)}/M$.
        \EndFor
        \State $E'^Q_{val} \leftarrow \sum_{j \in I_V}{(y(j)-Q(X_Q(j)|\theta^Q)^2}$.
        \If {$E'^Q_{val}<E^Q_{val}$}
        	\State $E^Q_{val} \leftarrow E'^Q_{val}$
        	\State $Q_{best}=\theta^{Q}$
        \EndIf
        \State $E'^{\mu}_{val} \leftarrow \sum_{j \in I_V}{Q([X_\mu(j),\mu(X_\mu(j)|\theta^\mu)]|\theta^Q)}$.
        \If {$E'^{\mu}_{val}<E^{\mu}_{val}$}
        	\State $E^{\mu}_{val} \leftarrow E'^{\mu}_{val}$
        	\State $\mu_{best}=\theta^{\mu}$
        \EndIf
    \EndFor
    \State \emph{Return}: The optimal $Q$ and $\mu$ neural network with weight parameter settings $Q_{best}$ and $\mu_{best}$ respectively.
\end{algorithmic}
\end{algorithm}

\subsection{Neural Network Design}
\label{stn:nn_design}

\begin{figure}[!t]
\centering
\includegraphics[width=0.7\columnwidth] {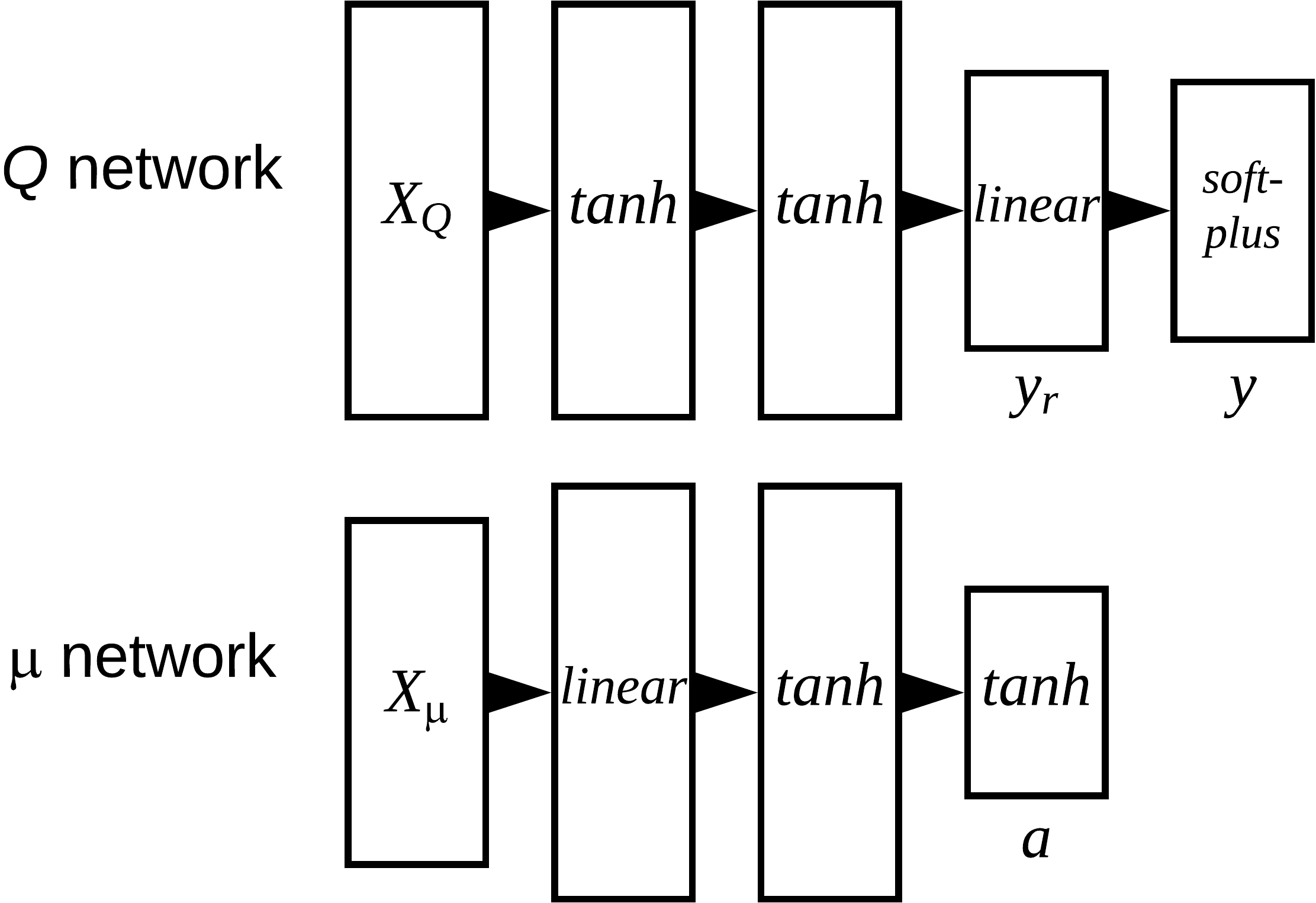}
\caption{Architecture of the $Q$ and $\mu$ network. The design follows the standard actor-critic architecture. One thing notable is that the second last layer of $Q$ outputs $y_r$, which can be used to compute the loss $y$ according to \eqref{eq:opt}.}

\label{fig:nn_design}
\end{figure}

The setup of the proposed $Q$ and $\mu$ network is shown in Fig. \ref{fig:nn_design}. For the $Q$ network, it has three hidden layers (two layers with $tanh$ activation and one with linear output) and outputs the negative reward $y$. To reduce the learning difficulty, the second last layer of $Q$ outputs the predicted energy and temperature data, concatenated as $y_r$, which is then used to compute $y$ according to \eqref{eq:opt}. In training, $Q$ is trained by minimizing the error between the predicted $y_r$ and the real data. For the $\mu$ network, it has two hidden layers (one with linear activation and one with $tanh$ activation function) and outputs the next control action $a$. $\mu$ network is optimized to reduce the loss function computed by the $Q$ network. We found that a variety of different neural network architectures can achieve similar results with necessary hyper-parameter tuning. In Section \ref{stn:test_nn_design} we show the experimental results on comparing different network architectures. Note that to fit for the $tanh$ activation used in the neural networks, we normalize all data entries into the range (-1, 1) and de-normalize the output of the neural networks when they are needed to compute the real energy and temperature values.

\section{Simulation Based Numerical Evaluation and Analysis} 
\label{stn:test_sim}
In this section, we present numerical evaluation results of the proposed CCA based on simulation. Simulation on the EnergyPlus platform is carried out to collect the training data and evaluate the proposed algorithms. Two baseline algorithms are compared with our proposed solution: one is the default control algorithm DefaultE+ from EnergyPlus, which computes the set-points according a target zone temperature with the underlying model; another is a general TS control optimization algorithm adapted from \cite{chow2002global} which is trained with the same data for the proposed approach.

\subsection{Simulation Configurations}  \label{SimConf}
We use EnergyPlus to collect the training data and assess different control algorithms for the following reasons. First, it is impossible to directly test control algorithms on a real DC due to the potential risk and the long running time. Second, EnergyPlus, whose development is an initiative of the U.S. Department of Energy Building Technologies Office, is a widely recognized and reliable simulation platform to model building cooling energy consumption. Third, EnergyPlus provides the flexibility that allows simulation with user-defined algorithms, control actions, and schedules.

The simulation is configured as follows. We adopt the original DC model provided by the EnergyPlus platform to make this simulation-based study tractable, as shown in Section \ref{model}.  We choose Singapore as the location and select the corresponding weather file to revise the simulation configuration file accordingly. We use a CPU loading trace collected from the monitoring system of Wikimedia as the workload trace for the DC model in the simulation. The whole simulation period is one year, and simulation data are collected every 6 minutes.

We use a random control algorithm to generate a one-year simulation trace to train the proposed algorithm. The control variables are randomly selected from valid ranges (obtained from a trace generated by the DefaultE+ algorithm) for the simulation model and then smoothed to ensure that the actions fluctuate smoothly. For the whole one-year simulation period, we select the last 45\% as the test period.

\subsection{Algorithm Configurations} \label{AlgoConf}
For the proposed CCA, the hidden layer sizes of $Q$ network are set to 50, 50 and 3, and 50, 50 for the $\mu$ network. The optimization algorithm to update the weight parameters is Adadelta \cite{zeiler2012adadelta}. The maximum training epoch $MaxEpoch$ is set to 200. The training batch size is set to 128. The penalty factor $\lambda$ in \eqref{eq:opt} is set to 0.01; as $\lambda$ directly controls the trade-off between energy cost and cooling effects, we need to manually tune this parameter. More on settings of $\lambda$ are shown in Section IV-E. The temperature threshold $\phi$ in \eqref{eq:opt} is set to 29. Setting of $\phi$ is depending on the target temperature one want to achieve.

For the TS optimization algorithm, we adopt the approach from \cite{chow2002global} with the following changes otherwise it will be incomparable to our approach. In the first stage, we train the same evaluation network like CCA to replace the original neural network designed for modeling chiller efficiency in \cite{chow2002global}. In the second stage, an iterative differential evolution (DE) optimization algorithm provided by Scipy \cite{jones2001open} is used to find the optimal solution for each test state. Ideally, the TS algorithm can perform no worse than the CCA algorithm if the optimization algorithm itself is optimized for this problem. Here as we focus on the design of CCA, for TS approach, a general optimization algorithm is used like in \cite{chow2002global}; this is reasonable as in the real case designing a special optimization algorithm is not an easy task.

For the proposed CCA and the TS optimization algorithms, the optimal control settings generated by these two algorithms are tested by simulation on the EnergyPlus platform. That is, for each state at the testing phase, we use the settings provided by the CCA or the TS algorithm and then record the resulting state changes and rewards for performance evaluation.

\subsection{Comparing CCA to Baseline Algorithms}
In this section, we present the simulation results of the average PUE and maximum outlet temperature (during the test period), obtained by using DefaultE+, TS, and CCA algorithms. Based on these results we further compare and evaluate the underlying control algorithms.

\begin{table}[]
\centering
\caption{Comparison of the proposed CCA with the baseline algorithms: DefaultE+ and TS. The results are based on 10 times of independent tests.}
\label{tbl:comp2baseline}
\resizebox{\columnwidth}{!}{%
\begin{tabular}{c|c|c|c|c|c}
\hline
\multicolumn{1}{l|}{} & DefaultE+ & TS $\tau=1$ & \textbf{CCA $\tau=1$} & TS $\tau=3$ & CCA $\tau=3$ \\ \hline
PUE & 1.376$\pm$0.000 & 1.371$\pm$0.000 & \textbf{1.333$\pm$0.003} & 1.346$\pm$0.023 & 1.307$\pm$0.017 \\ \hline
$max(T_{z_1})$ & 28.617$\pm$0.000 & 26.242$\pm$0.000 & \textbf{28.910$\pm$0.361} & 30.995$\pm$3.121 & 30.111$\pm$2.759 \\ \hline
$max(T_{z_2})$ & 28.655$\pm$0.000 & 26.288$\pm$0.000 & \textbf{29.124$\pm$0.353} & 30.291$\pm$3.307 & 33.250$\pm$0.005 \\ \hline
\end{tabular}%
}
\end{table}

\begin{figure}[]
\centering
\includegraphics[width=0.95\columnwidth] {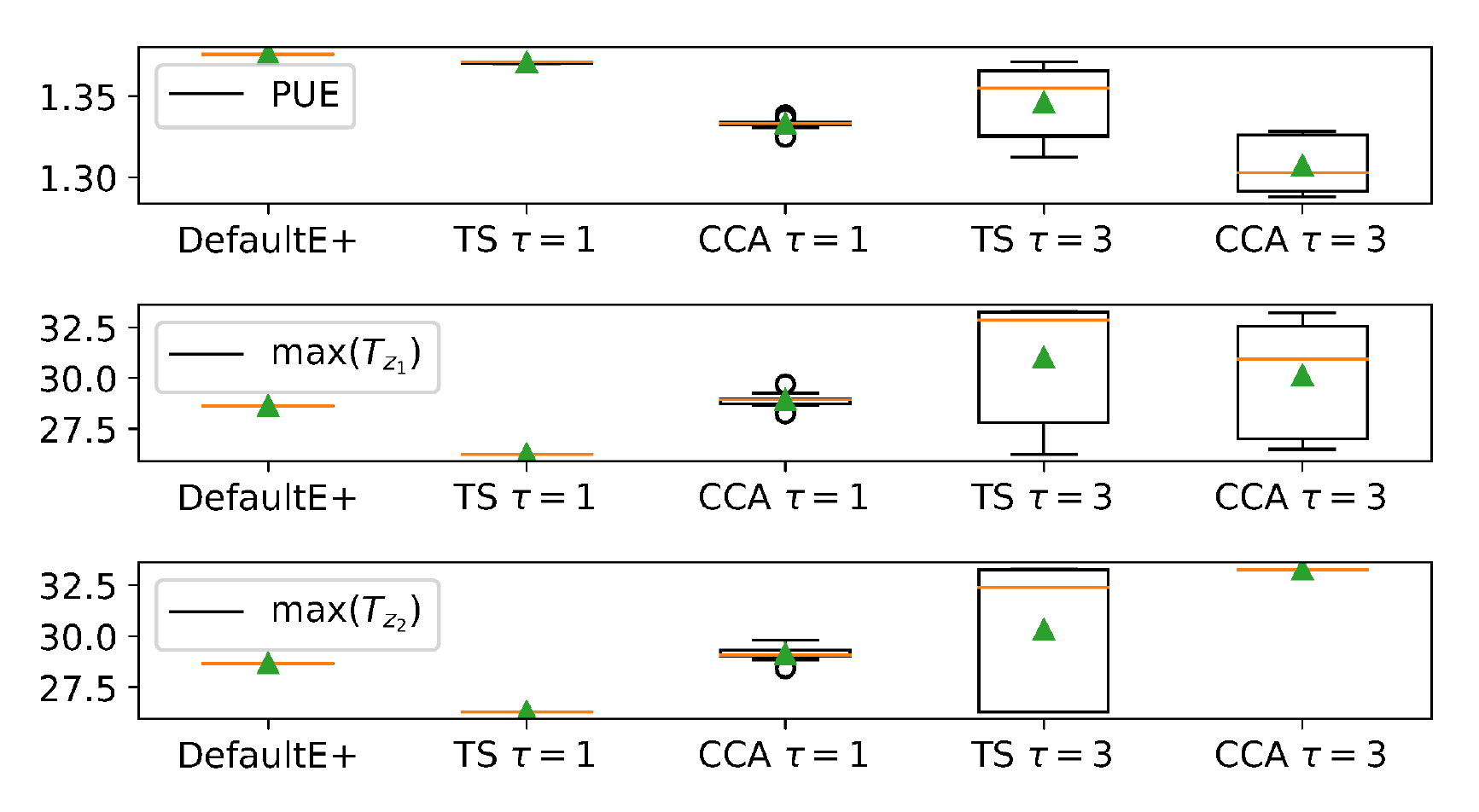}
\caption{Distribution of PUE and maximum outlet temperature $T_{z_1}$ and $T_{z_2}$ over the test period achieved by different control algorithms, respectively. The results are based on 10 independent tests. The orange line in each box is the median, and the green triangle is the mean value.}

\label{fig:boxerr}
\end{figure}

Table \ref{tbl:comp2baseline} shows the first and second-order statistics of the PUE and maximum outlet temperature in 10 independent runs with different $\tau$ settings for TS and CCA. For better examining these results, we also plot the data distribution in Fig. \ref{fig:boxerr}. We can observe the following:
\begin{itemize}
\item The proposed CCA algorithm with $\tau=1$ achieves the best control results, by reducing the PUE from 1.37 to 1.33, which is 11\% cooling power saving, while maintaining the temperature of both zones under or nearby the pre-defined threshold 29. This shows that the actor network can indeed attain optimal or close-to-optimal control settings. 
\item The TS algorithm with the general optimization algorithm shows unstable performance. To improve its performance, a specialized optimization procedure will be necessary. Being compared to the TS approach, the proposed CCA is an end-to-end solution. CCA can directly output the control setting with the pre-trained policy network, which can be carefully tuned and tested in an offline manner; while for the case of TS, an online optimization algorithm has to be used, which poses higher computation cost and accuracy problem in real time.
\item The recurrent version of CCA with $\tau=3$ has unstable results. This is reasonable as recurrent decision making can be beneficial if the data are noisy; however, as a simulated case is studied here, the data generated are free of noise. In Section \ref{stn:test_NSCC} we show how recurrent decision making can be useful in a case with real data collected from a physical DC.
\end{itemize}

\begin{figure}[!t]
\centering
 \subfigure[]{
   \includegraphics[width=0.95\columnwidth] {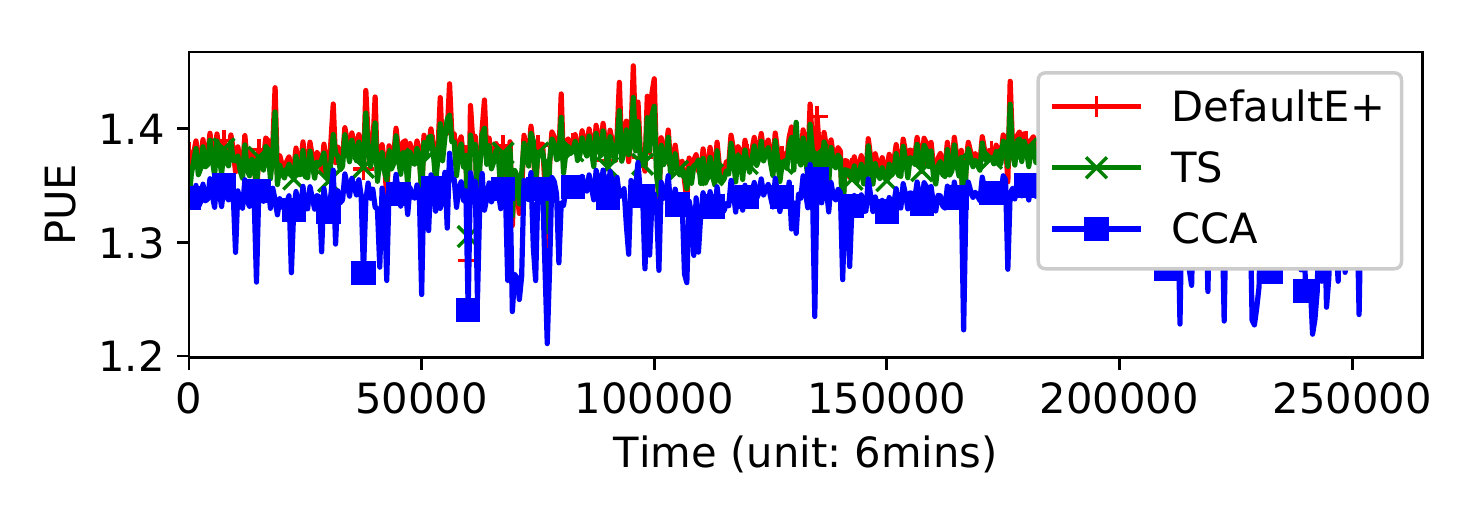}
 }
 \subfigure[]{
   \includegraphics[width=0.95\columnwidth] {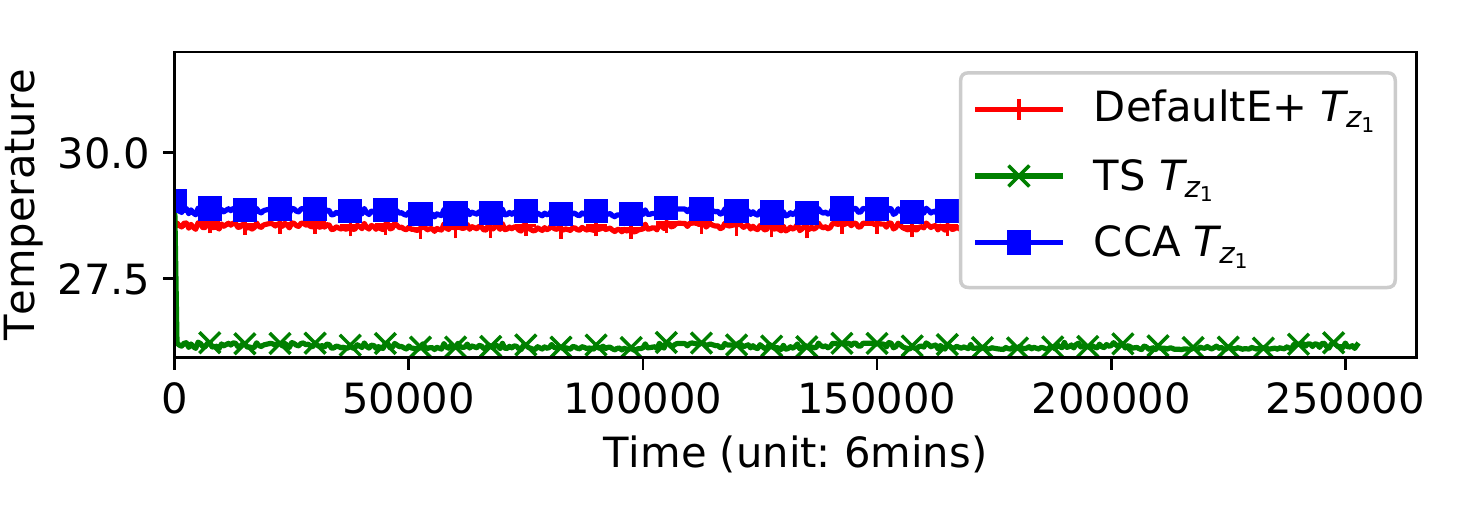}
 }
 \subfigure[]{
   \includegraphics[width=0.95\columnwidth] {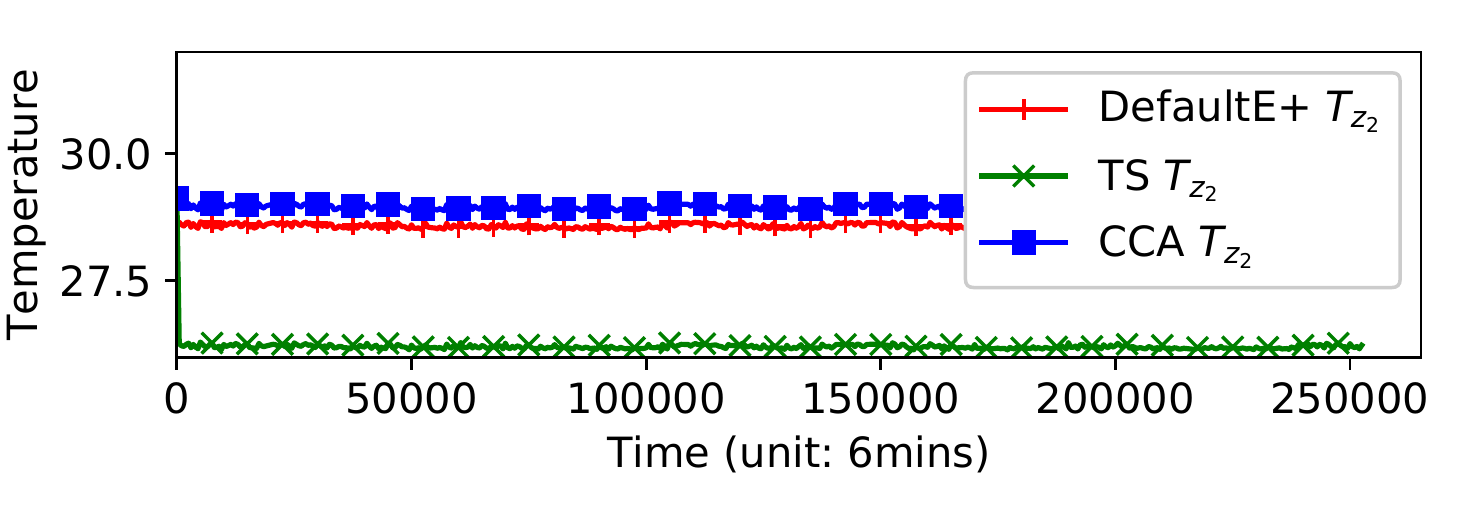}
 }
\caption{Simulation example comparison of DefaultE+, TS and CCA. (a) PUE; (b) $T_{z_1}$; and (c) $T_{z_2}$. All temperatures are in $^{o}$C. }
\label{fig:simre}
\end{figure}

We bring an example of the PUE and temperature traces (during the test period) obtained from our simulation in Fig. \ref{fig:simre}. We can point out that the PUE curve of CCA is lower than that of DefaultE+ and TS while achieving higher but still satisfying temperature curves in both zones. Note that at the beginning of the test period, the temperature of the DC zone has a fast drop for TS, due to transient from the DefaultE+ algorithm to the learning algorithm (the DefaultE+ algorithm provides the settings before the test period).

\subsection{Neural Network Design Study}
\label{stn:test_nn_design}
In this subsection, we study different designs of the neural network and compare their performances.  We compare our network design with three other different implementations: 1) TargetNet: with the target network of DDPG, which is used to avoid fast changing of the $Q$ network to stabilize the training; 2) ReluNet: a four-relu-layer based $Q$ network (1024-512-256-3) similar to \cite{newarc} to test whether relu activation can work or not;  3) LstmNet: with a LSTM layer to process the recent history trace. As we use the recent states and actions in the last $\tau$ steps as the input to the $Q$ network, we can use an LSTM layer to process it first as LSTM is a recurrent neural network which is suitable for dealing with sequential data. The LSTM layer outputs its hidden units which are then fed into a normal $Q$ network as described in Section \ref{stn:nn_design}. We test these different designs with $\tau$ set to 1; for LstmNet, we show the results with $\tau$ set to 3.

Test results of these different designs are shown in Table \ref{tbl:network}. Comparing to the original CCA, Table \ref{tbl:network} shows that these different architectures achieve very similar results. With TargetNet, the results are almost the same to the original design, which is reasonable as we use a long offline trace to train the $Q$ network in each training episode, which can lead to a very stable learning process. With relu layers, we achieve similar results but slightly higher temperature reading with lower PUE; with LSTM layer added, we achieve a slightly lower temperature but higher PUE. Given sufficient hyper-parameter tuning, these designs may also achieve satisfying performance. In this simulation case, as adding larger size relu layers and LSTM layers can slow down the training speed, we adopt the simpler design illustrated in Fig. \ref{fig:nn_design}.

\begin{table}[]
\centering
\caption{Comparing different neural network implementations: TargetNet, ReluNet, LstmNet with the original CCA design. All temperatures are in $^{o}$C. }
\label{tbl:network}

\resizebox{\columnwidth}{!}{%
\begin{tabular}{c|c|c|c|c}
\hline
\multicolumn{1}{l|}{} & CCA & TargetNet & ReluNet & LstmNet \\ \hline
PUE & 1.333e+00$\pm$3.371e-03 & 1.338e+00$\pm$1.301e-03 & 1.326e+00$\pm$2.128e-07 & 1.343e+00$\pm$2.305e-02 \\ \hline
$max(T_{z_1})$ & 2.891e+01$\pm$3.614e-01 & 2.850e+01$\pm$1.281e-01 & 2.949e+01$\pm$7.958e-14 & 2.874e+01$\pm$2.186e+00 \\ \hline
$max(T_{z_2})$ & 2.912e+01$\pm$3.525e-01 & 2.867e+01$\pm$1.208e-01 & 2.951e+01$\pm$1.709e-09 & 2.864e+01$\pm$2.640e+00 \\ \hline
\end{tabular}%
}
\end{table}

\subsection{Hyper-Parameter Setting Study}
\label{para_study}
Despite network architecture studied above, the hyper-parameter setting is also a critical issue in applying DRL algorithm to practical applications. In this subsection, we discuss how we set the hyper-parameters used in our algorithm, such that it may shed some light on similar applications. A key hyper-parameter is the learning rate. In our approach, we choose to use Adadelta as the training optimization algorithm, which does not oblige us to set the learning rate manually. In our case, Adadelta works great, yet it may not work on other problems. When the learning rate needs to be set manually, we recommend trying from a small value like 1e-5. Another key hyper-parameter is the initialization range of the weight of the neural networks. We use zero mean and 0.01 as the standard deviation to generate the random weights. Smaller initialization range tends to result in a more stable training process.

\begin{figure}[!t]
\centering
   \includegraphics[width=0.95\columnwidth] {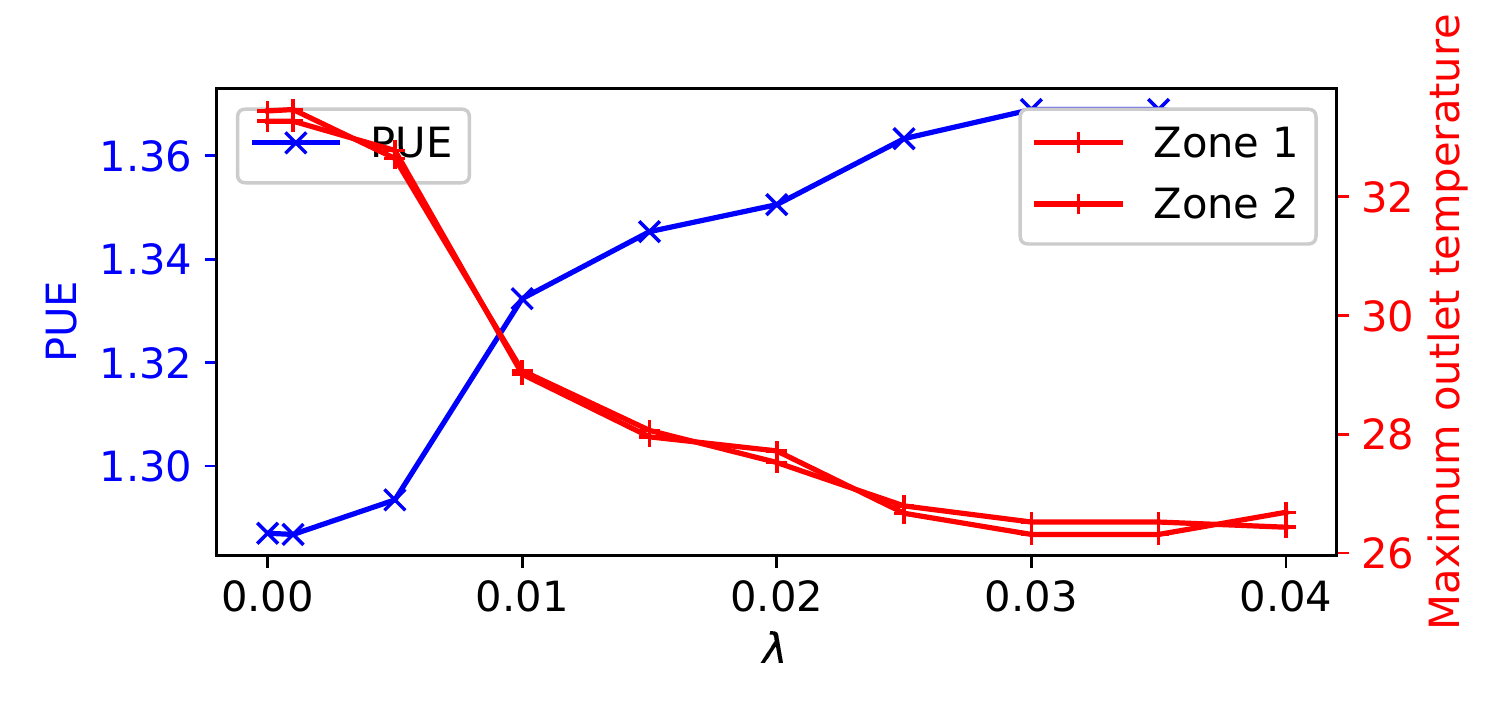}
\caption{Testing different settings for $\lambda$ from 0.0 to 0.04.}
\label{fig:lambda_trade}
\end{figure}

A critical hyper-parameter that related to the loss function in CCA is the penalty factor $\lambda$ in \eqref{eq:opt}. In our experiments, we found that it needs to be manually tuned to see a satisfying trade-off between energy and temperature. In Fig. \ref{fig:lambda_trade} we show when we change $\lambda$ from 0.0 to 0.04, how PUE and the maximum outlet temperature of each zone changes accordingly. We notice that a proper setting of $\lambda$ will be distinct from problem to problem. For example, in the next section, we will show the result of a different test case, in which the best $\lambda$ setting is different from this simulation case.

\section{Tests on Real Data Trace from NSCC}
\label{stn:test_NSCC}

\begin{figure}[!t]
\centering
\includegraphics[width=0.7\columnwidth] {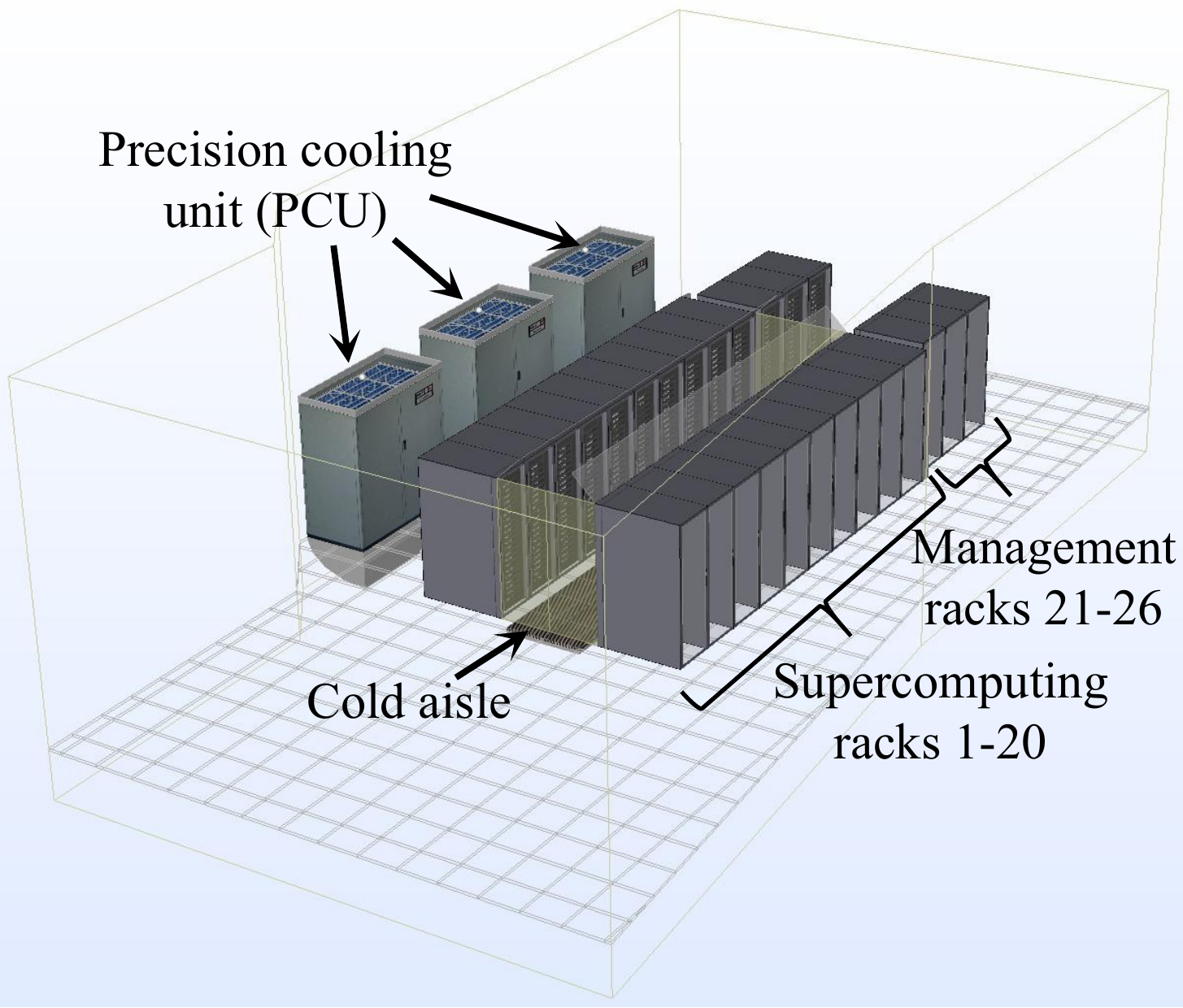}
\caption{3-D model of the target system in NSCC. There are 26 racks, which are cooled by the PCUs. The cold air supplied by the PCUs first goes through the raised floor to the cold aisle, and then cools the racks before returning to the PCUs.}
\label{fig:NSCC_3d}
\end{figure}

To further investigate the proposed algorithm, we test the proposed CCA on a data trace collected from the National Super Computing Centre (NSCC) of Singapore and show its performance on optimizing the energy cost while satisfying the cooling requirements (rack intake temperature). 

We are focused on the optimization of the air cooling system for the computing nodes in NSCC. The 3-D model of the research target is illustrated in Fig. \ref{fig:NSCC_3d}. There are 26 racks in the target system. Three precision cooling units (PCUs) supply cold air for these racks. The PCUs supply cold air at about 20 degrees. Cold air enters the cold aisle and then goes through the racks and at last returns back to the PCUs. There are other cooling facilities installed: for racks 1-20, an additional warm water cooling system is used to cool the CPU/GPU and memory chips; for racks 21-26, an additional rear-door cooling system is used. The warm water cooling system and the rear-door cooling system will not be studied here thus we omit further details.

We try to optimize the total supply flow rate of the target PCUs shown in Fig. \ref{fig:NSCC_3d}, aiming to minimize their power consumption while maintaining the average intake temperature of the racks (measurement point at the height of the 36U of each rack). 

To apply the proposed algorithm, we collect the related data for the optimization goal, as shown in Table \ref{tbl:NSCC_data}. Note that several measurements of the warm water cooling system and rear-door cooling system are also used. Our experiments show that including these reading can increase the approximation accuracy of the $Q$-network. We collected these data entries for every 3 minutes from March 1 to 15 of 2017. For these data, we use the first 85\% as the training data and the last 15\% as the test data. With the above data, we utilize the proposed algorithm to train the $Q$ and $\mu$ network. In this case, as the power consumption can be directly computed by the fan law from the airflow rate, we will only rely on the $Q$ to approximate the inlet temperature which we use as the thermal indicator.

\begin{figure}[!t]
\centering
\includegraphics[width=0.9\columnwidth] {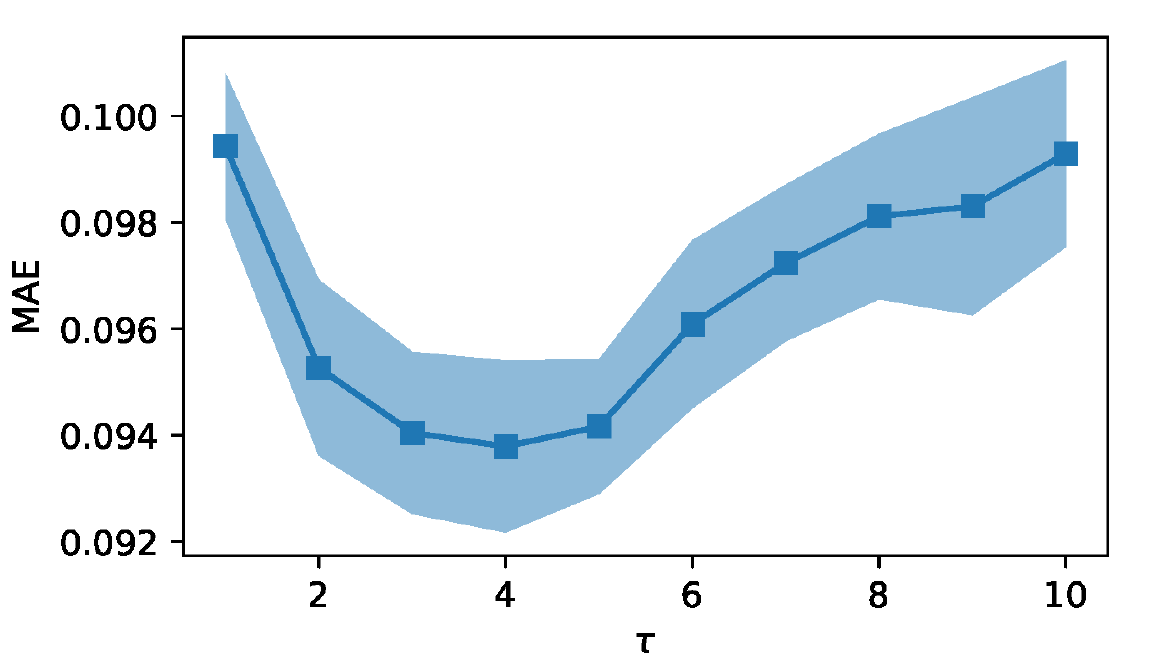}
\caption{Testing different $\tau$ settings from 1 to 10 for the NSCC data trace. Results show that with $\tau$=4, the best $Q$ learning quality can be achieved with the lowest MAE around 0.094 for the temperature prediction.}
\label{fig:tau_test}
\end{figure}

\begin{figure}[!t]
\centering
\includegraphics[width=0.9\columnwidth] {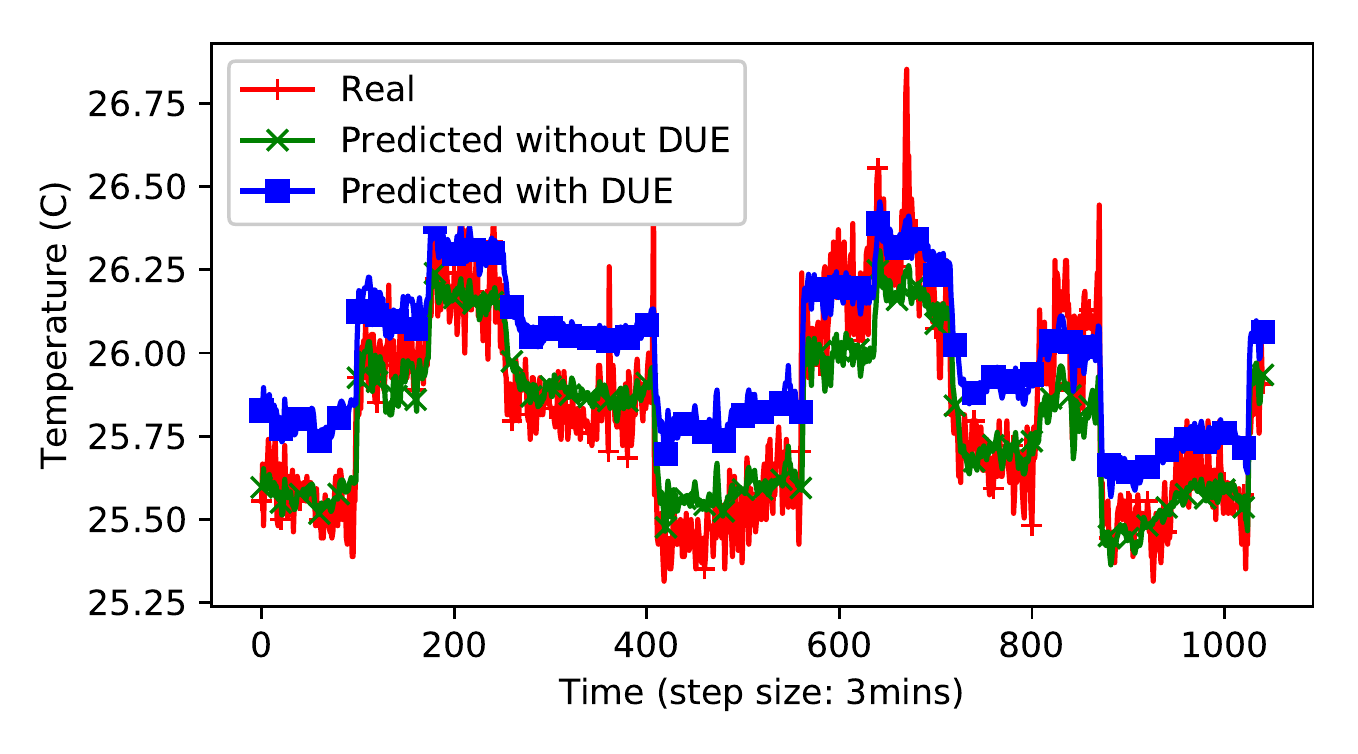}
\caption{Comparing the estimated temperature curve with and without DUE. We can observe that with DUE, although the prediction error is increased, the cases when the network underestimates the temperature are largely resolved.}
\label{fig:DUE_test}
\end{figure}

Experimental results demonstrate that the proposed algorithm works well as expected. First, the results show that the normalized mean absolute error (MAE) of the $Q$ network is smaller than 0.1 degrees, as shown in Fig. \ref{fig:tau_test}. This proves that the network can successfully capture the system dynamics. Fig. \ref{fig:tau_test} also shows that when dealing with real data with noise, recurrent decision making is better as we found that when setting $\tau=4$ we can get the best temperature estimation results. Even though MAE lower than 0.1 degree seems great, it can underestimate the temperature as the regression process will try to fit more on most of the data falling in the middle of the distribution. This may lead to over-optimistic energy saving estimation. As we cannot directly apply the CCA algorithm to a real DC, it is important that we can generate some convincing theoretical results first without underestimation. To solve this problem, we change the validation strategy in Algorithm \ref{alg:CCA} and apply a special de-underestimation (DUE) validation method, which works in the following manner. In line 11 of CCA, when computing the validation error, we replace the original square error into a function which only considers the underestimation error, as shown below:

\begin{equation}
E'^Q_{val} \leftarrow \sum_{j \in I_V}{max(y(j)-Q(X_Q(j)|\theta^Q, 0)}.
\end{equation}
With such DUE validation method, we can reduce the underestimation cases, as shown in Fig. \ref{fig:DUE_test}.

\begin{figure}[!t]
\centering
\includegraphics[width=0.9\columnwidth] {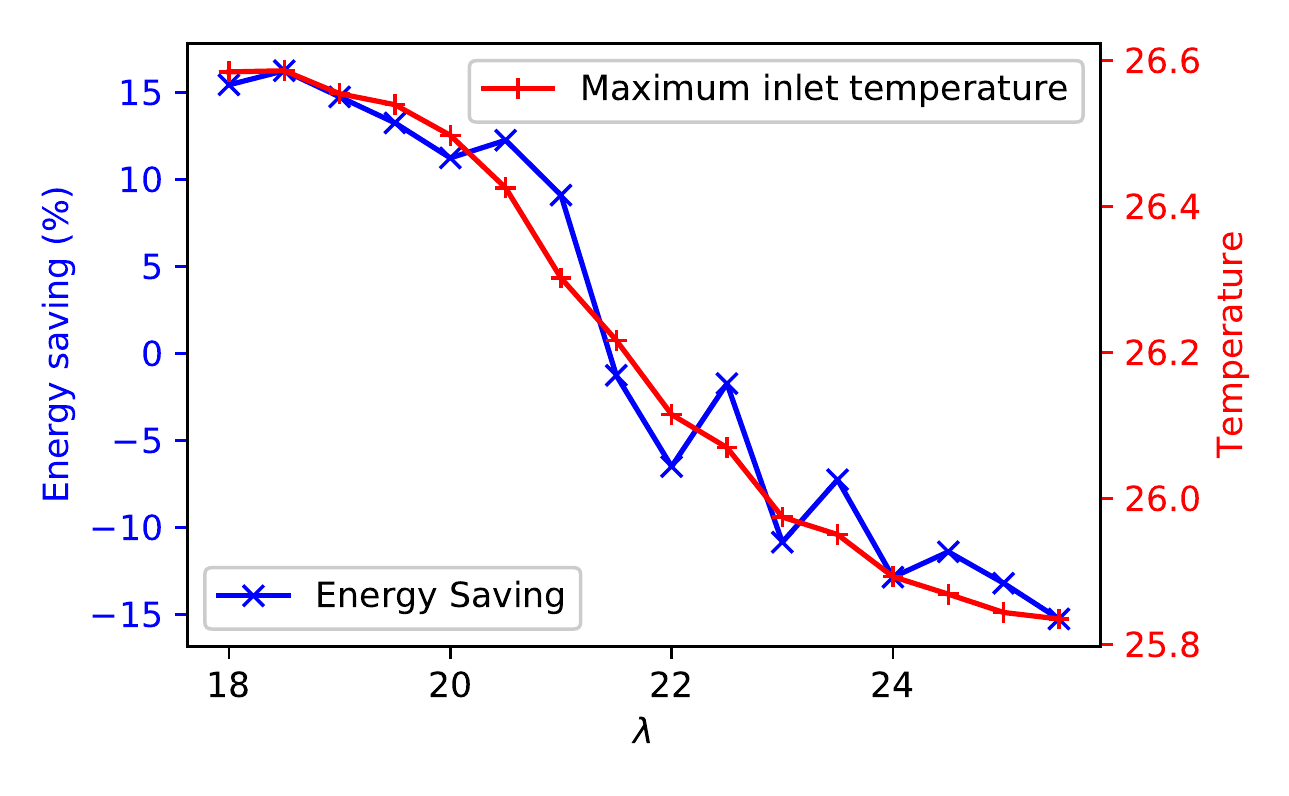}
\caption{Testing different $\lambda$ settings for the NSCC case. Energy saving decreases when $\lambda$ increases, at the same time the maximum temperature decreases.}
\label{fig:lambda_test_NSCC} 
\end{figure}

\begin{figure}[!t]
\centering
 \subfigure[]{
   \includegraphics[width=0.95\columnwidth] {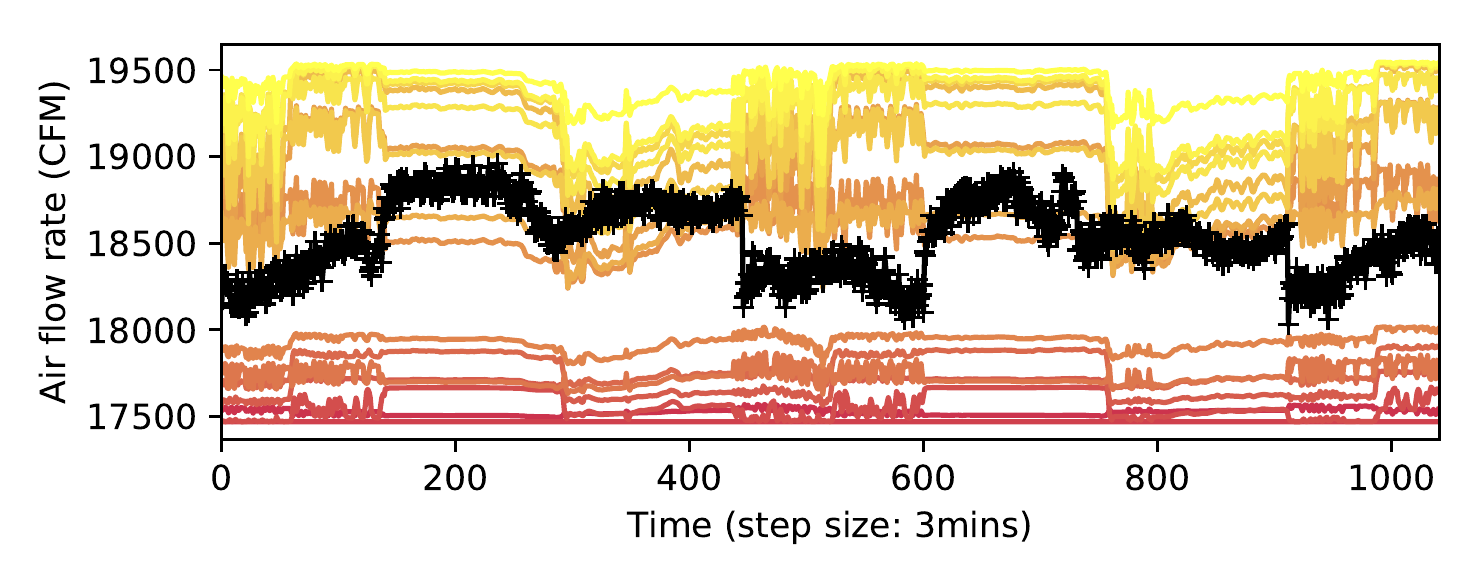}
 }
 \subfigure[]{
   \includegraphics[width=0.95\columnwidth] {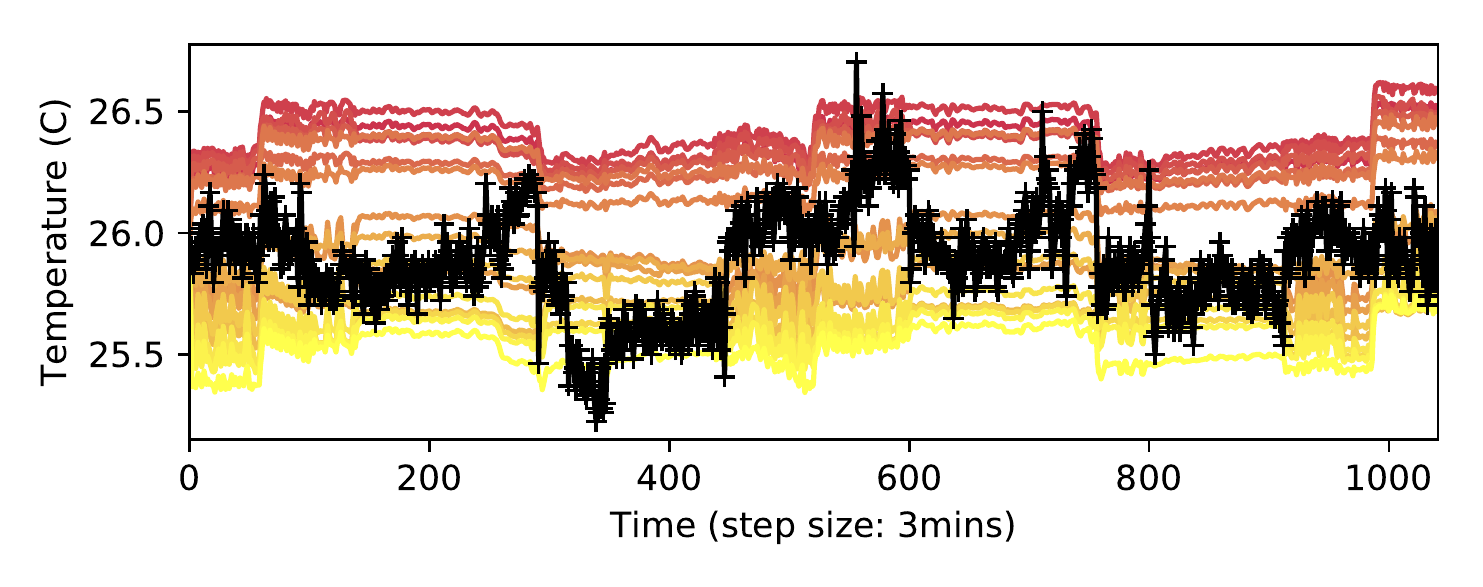}
 }

\caption{Test results on the NSCC data trace when $\lambda$ changed from 18 to 26: (a) control setting (airflow rate) predicted by the proposed algorithm; (b) the correspondingly predicted rack temperature. Darker color curves are with smaller $\lambda$ settings, and the black curve (with "+" marker) is the real data trace collected in the test period.}

\label{fig:ctrset}
\end{figure}

Second, we study how much energy saving we can achieve when using different penalty parameter $\lambda$ (used in the objective function \eqref{eq:opt}), as shown in Fig. \ref{fig:lambda_test_NSCC}. In Fig. \ref{fig:ctrset} we also show how the control actions and predicted temperature change with different $\lambda$. From Fig. \ref{fig:lambda_test_NSCC} we can see that with DUE, we can achieve energy saving 15\% easily if we set the temperature threshold larger than 26.6 degrees. If we want a lower maximum temperature such as 26.4, we can still save about 10\%. When we set the target maximum temperature lower than 26.2, we will have to use more cooling power than the actual setting adopted. 

\begin{table}[]
\centering
\caption{The data entries and their usage in optimization of the airflow rate of the target PCUs.}
\label{tbl:NSCC_data}
\resizebox{0.95\columnwidth}{!}{%
\begin{tabular}{c|l}
\hline
State & \begin{tabular}[c]{@{}l@{}}P1: total power consumption of racks 1-26; \\ P2: total power consumption of the other racks; \\ P3: total heat load recorded in warm water cooling system; \\ F1: water flow rate; T1: water flow supply temperature; \\ F2: air flow rate of other PCUs; \\ T2: average supply temperature of other PCUs; \\ F3: rear door cooling distribution units chilled water flow rate; and \\ T3: the average supply temperature of the target PCUs.\end{tabular} \\ \hline
Action & F4: the airflow supply rate of the target PCUs \\ \hline
Reward & \begin{tabular}[c]{@{}l@{}}P4: power consumption of the target PCUs; and\\ T4: average rack intake temperature.\end{tabular} \\ \hline
\end{tabular}%
}
\end{table}   

\section{Conclusion}
\label{stn:summary}

DC powers the modern society as the infrastructure for information storage, processing, and dissemination. At the same time, DC consumes a formidable amount of electricity, among which a significant portion is used in cooling. To develop an optimal control policy for the complex cooling system for a DC is a complex task. We propose and verify an end-to-end DRL approach CCA for the control optimization of the cooling system of a DC. Compared to the existed TS optimization method, the proposed CCA can directly optimize a policy network based on the observed historical data, while the policy can output the optimized control settings for any given state. Adapted from DDPG and the actor-critic framework, our algorithm is a batch off-policy algorithm, with which we tested various algorithm settings to examine performance thoroughly.

We evaluate the proposed algorithm on the simulation platform EnergyPlus and a trace collected from a real DC. The simulation results show that our method can maintain the DC temperature within the predefined threshold while achieving lower PUE, and save about 11\% cooling energy compared to a baseline approach with manually designed control settings. Our results on the real trace show that we can achieve high evaluation accuracy and the predicted control setting can reduce the cooling cost around 15\% while maintaining the rack intake temperature under a predefined threshold. The results prove that our algorithm can successfully learn the system dynamics from the monitoring data and can contribute to improving the cooling efficiency.




%

\bibliographystyle{IEEEtran}
\bibliography{IEEEabrv,thermal_optimization}

\newpage

\end{document}